\documentclass[10pt,twocolumn,letterpaper]{article}

\usepackage{cvpr}
\usepackage{times}
\usepackage{epsfig}
\usepackage{graphicx}
\usepackage{amsmath}
\usepackage{amssymb}
\usepackage{subcaption}
\usepackage{float}
\usepackage[english]{babel} 

\usepackage[T1]{fontenc}
\usepackage{babel}
\usepackage[font=small,labelfont=bf,tableposition=top,skip=3pt]{caption}
\usepackage{blindtext}
\usepackage{cuted}
\usepackage{enumitem}
\usepackage[utf8]{inputenc}
\usepackage{authblk}

\graphicspath{ {./images/} }

\usepackage[breaklinks=true,bookmarks=false]{hyperref}

\cvprfinalcopy 

\def\cvprPaperID{****} 

\begin{document}

\title{UrbanLoco: A Full Sensor Suite Dataset for Mapping and Localization in Urban Scenes}

\author[1]{Weisong Wen\thanks{Both authors contributed equally to this work.}}
\author[2]{Yiyang Zhou$^*$}
\author[1]{Guohao Zhang}
\author[2]{Saman Fahandezh-Saadi}
\author[1]{Xiwei Bai}
\author[2]{Wei Zhan}
\author[2]{Masayoshi Tomizuka}
\author[1]{Li-Ta Hsu\thanks{lt.hsu@polyu.edu.hk}}
\affil[1]{Intelligent Positioning and Navigation Lab, Hong Kong Polytechnic University}
\affil[2]{Mechanical Systems Control Lab, University of California, Berkeley}
\maketitle
\noindent
 
\begin{strip}
\vspace{-25mm}
\centering\noindent
\begin{center}
\includegraphics[width=1.0\textwidth]{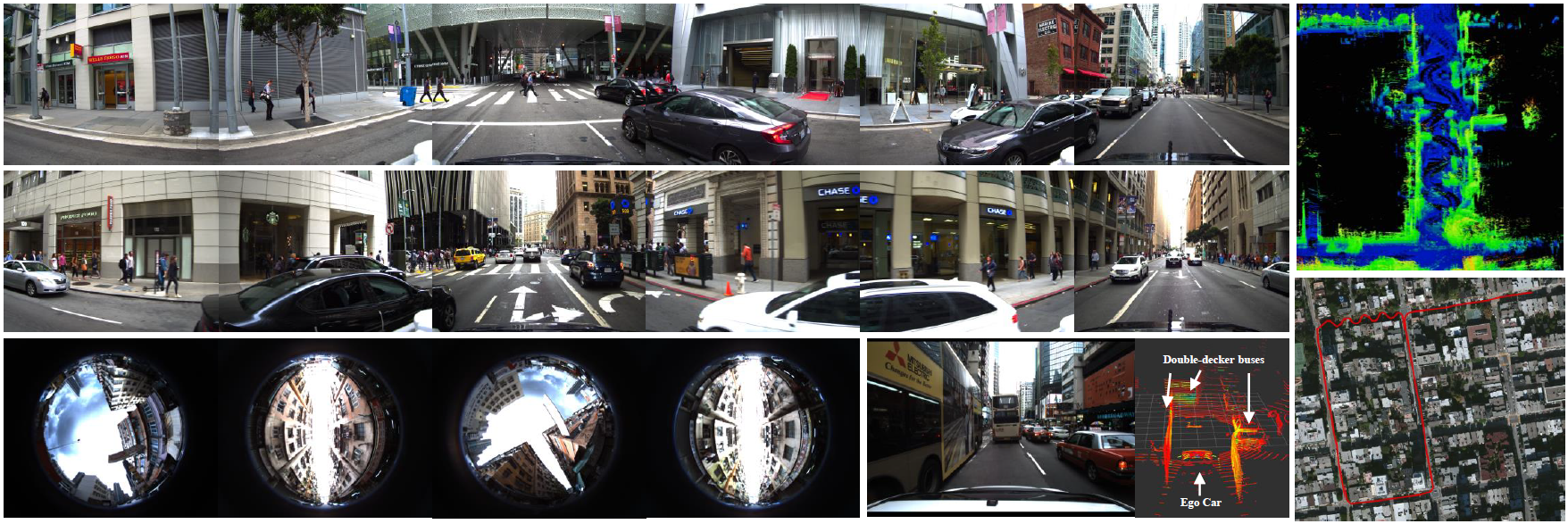}
\captionof{figure}{An Overview of the UrbanLoco Dataset: The UrbanLoco dataset focuses on highly urbanized areas in San Francisco and Hong Kong with a full sensor suite: 360 degree camera (San Francisco), fish eye sky camera (Hong Kong), LIDAR, GNSS receivers and IMU. The dataset covers various road conditions including tunnels, urban canyons, construction sites, sharp maneuvers, hills, etc.}
\end{center}
\end{strip}

\begin{abstract}
   Mapping and localization is a critical module of autonomous driving, and significant achievements have been reached in this field. Beyond Global Navigation Satellite System (GNSS), research in point cloud registration, visual feature matching, and inertia navigation has greatly enhanced the accuracy and robustness of mapping and localization in different scenarios. However, highly urbanized scenes are still challenging: LIDAR- and camera-based methods perform poorly with numerous dynamic objects; the GNSS-based solutions experience signal loss and multipath problems; the inertia measurement units (IMU) suffer from drifting. Unfortunately, current public datasets either do not adequately address this urban challenge or do not provide enough sensor information related to mapping and localization. Here we present UrbanLoco: a mapping/localization dataset collected in highly-urbanized environments with a full sensor-suite. The dataset includes 13 trajectories collected in San Francisco and Hong Kong, covering a total length of over 40 kilometers. Our dataset includes a wide variety of urban terrains: urban canyons, bridges, tunnels, sharp turns, etc. More importantly, our dataset includes information from LIDAR, cameras, IMU, and GNSS receivers. Now the dataset is publicly available through the link in the footnote \footnote[1]{Dataset Link: \url{https://advdataset2019.wixsite.com/urbanloco}}.

\end{abstract}

\begin{table*}[ht!]
\begin{center}
\caption{Datasets Comparison}
\vspace{2mm}
\begin{tabular}{|c|c|c|c|c|c|c|c|}
\hline
DATASET& LOCATION & URBANIZATION  & DISTANCE                   & LIDAR & CAMERA & IMU & GNSS   \\
 NAME  &          &   RATE        &(Max Path) &       &        &     &        \\
\hline
\hline
KITTI\cite{KITTI} & Karslruhe & Low & 1km & YES& front & YES & NO  \\
RobotCar\cite{RobotCarDatasetIJRR} & Oxford & Low & 1km & YES& front & YES & NO  \\
KAIST\cite{kaist} & Seoul & High & 11.42km & Tilted& front & YES & YES  \\
NuScenes\cite{nuScenes} & Singapore, Boston & High & 20s & YES& 360  & YES & YES  \\
Waymo\cite{waymo_open_dataset} & SF Bay Area & High & 20s & YES& 360  & NO & NO  \\
Lyft\cite{lyft2019} & San Francisco & High & 20s & YES& 360  & NO & NO  \\
Argoverse\cite{Argoverse} & Miami, Pittsburgh & Mid & 30s & YES& 360  & NO & NO  \\
\hline
Ours & San Francisco, Hong Kong & High & 13.8 km & YES& 360  & YES & YES  \\
\hline
\end{tabular}
\end{center}
\vspace{-5mm}
\end{table*}

\section{Introduction}
Publicly available datasets have been hugely influential in autonomous driving research, both in academia and in industries. In the past few years, numerous datasets were published to advance different aspects of autonomous driving. One of the pioneers in academia is KITTI \cite{KITTI}, which covered most topics in current research and launched benchmarks for evaluation. Most recently, a few companies, including Lyft \cite{lyft2019}, nuTonomy \cite{nuScenes}, Waymo \cite{waymo_open_dataset}, and Argo AI \cite{Argoverse}, published their datasets for detection and prediction. However, few of these publicly available datasets addresses the urban mapping and localization problem, which is yet to be solved with cost-effective sensors combination.

Due to high-rise structures and numerous dynamic objects, mapping and localization in a densely populated is hard, and simultaneous localization and mapping (SLAM)\cite{Grisetti} in a city is even harder. Figure 1 shows an overview of the city landscape in our dataset. The traditional Global Navigation Satellite System (GNSS) based solution fails in urban canyon due to limited satellite visibility and multipath problems. For more recent LIDAR or visual based approaches, the dynamic objects (vehicles, pedestrians, cyclists) may cause inaccurate point registration. While inertia measurement unit (IMU) is less affected by urban environments, the IMU suffers significantly from drifting over time. 

With the aforementioned challenges, however, most of the current public datasets are not specifically targeting at the mapping and localization tasks in dense urban applications. As densely populated areas are unavoidable for autonomous driving, it is urgent to provide the public with a mapping/localization-focused dataset in urban scenes. Here we present the UrbanLoco dataset: a full sensor suite dataset for mapping and localization in densely populated landscapes. The dataset includes information from 4 essential sensors: LIDAR, camera, IMU, and GNSS. The data was collected in the populous districts in Hong Kong and San Francisco. 

The major contributions of our dataset are:

\setlist{nolistsep}
\begin{itemize}[noitemsep]
    \item We are releasing the first large scale full sensor suite dataset focusing on urban mapping and localization challenges;
    \item The dataset includes over 40 kilometer trajectories covering various driving scenarios including urban canyons, bridges, hills tunnels, etc. There are also numerous dynamic objects in the scene;
    \item The dataset provides information from a full sensor-suite: 1 LIDAR, 6 cameras with 360 degrees view (1 camera in Hong Kong), 1 IMU, and 1 GNSS. The sensor-suite is carefully calibrated with calibration logs available online;
    \item We define the urbanization measure for localization in city landscapes and evaluated the urbanization rate on current mapping/localization datasets;
    \item The dataset is publicly available through the dataset website, and related APIs are also available for users' convenience.
\end{itemize}


\section{Related Works}

\textbf{Autonomous Driving Datasets for Mapping and Localization}
Mapping/localization-focused datasets and benchmarks have been heavily used by current researchers. KITTI \cite{KITTI} odometry is a popular benchmark collected in the German town of Karlsruhe. Most of the trajectories are long enough (>500 m) to test out-door localization algorithms. The total of 20 trajectories have been the popular test field for the latest algorithms. Indeed, KITTI odometry benchmark ranking is still updating nowadays. Unfortunately, KITTI dataset does not pose enough challenges for mapping and localization: the data was collected in rural areas with light traffic and relatively low-lying structures. Furthermore, KITTI only provides front-view cameras and LIDAR information, lacking the critical GNSS and IMU outputs. Oxford RobotCar \cite{RobotCarDatasetIJRR} is another popular dataset which specializes in mapping and localization. Unlike KITTI, the Oxford RobotCar dataset offers extra information from IMU and GPS measurements, which provides more possibilities for testings of various algorithm designs. However, the data is collected in the less-urbanized Oxford, not adequately addressing the urban mapping and localization problems. Table 1 shows a comparison of the aforementioned two datasets with our dataset. We would further quantify the urbanization rate specifically for localization in the next section.

\textbf{Autonomous Driving Datasets for Urban Scenes}
Until very recently, it was rare to see autonomous driving datasets in highly-urbanized scenarios. In early 2019, the Boston/Singapore based NuScenes dataset \cite{nuScenes} was published. Similar to our dataset, the NuScenes data is collected among dense traffics and high-rise structures. More importantly, NuScenes dataset also incorporates visual information and IMU. However, since the NuScenes dataset is majorly designed for perception and tracking, the data is segmented into scenes of 20 seconds in length. Thus, it is hard to acquire a sufficiently long trajectory for mapping and localization purposes. Later in 2019, Lyft \cite{lyft2019}, Waymo \cite{waymo_open_dataset}, and Argo AI \cite{Argoverse} published their dataset. These datasets also targets at detection and prediction as well; thus, they are similarly segmented into discrete pieces, which could not be used for long trajectory generation. Furthermore, these datasets did not include IMU and GPS information. Table 1 shows a general comparison of different public datasets for autonomous driving. 

{\bf GNSS Based Localization} GNSS is widely used as a convenient tool for localization tasks. Equipped with a GNSS receiver, the ego vehicle could easily navigate in regions where GNSS receptions and satellite visibility are satisfactory. However, such conditions are rarely met in urban scenarios. Indeed, GNSS manufactures are incorporating IMU with GNSS receivers to continuously navigate when the satellites are lost in sight. However, such equipment is usually expensive, and could not be used commercially for mass application. 

{\bf Vision-based Mapping and Localization} Stereo cameras were used to extract spatial information about a scene \cite{NisterVisualOdo} \cite{MaimomeVisualOdo}, but it is hard for monocular camera to solve the general motion. To complement the lack of depth information for monocular cameras, RGB-D cameras were utilized successfully for SLAMs in different scenes \cite{HuangRGBD} \cite{HenryRGBD} \cite{KerlRGBD}. However, the depth detection span is very limited compared to LIDARs. Another sensor fusion method for monocular navigation is the Visual Inertia Navigation System (VINS-MONO) \cite{qin2017vins}, which tightly couples a monocular camera along with an IMU for localization. A nonlinear optimization algorithm is applied to incorporate the loss in IMU measurement as well as the loss in visual feature matching. This specific algorithm is evaluated in Section 4.

{\bf Laser-based Mapping and Localization} Laser odometry is another significant branch of SLAM research. Point cloud registration algorithms are the cornerstones for laser-based methods: ICP and its multiple variations \cite{ICP} \cite{ICP-P2P} \cite{ICP-P2Plane} have been efficient algorithms in odometry estimation. More point registration algorithms like Normal Distribution Transformation (NDT) \cite{NDT} and Mix-Norm Matching (MiNoM) \cite{MiNoM} also improves the quality of pure laser odometry estimation. The LIDAR Odometry and Mapping Method (LOAM) \cite{LOAM} extracts surface and corner features in the scan to determine distances in a voxel grid-based map. In this paper, we present the evaluation of the performances of LOAM and NDT on our dataset. 

\section{Urbanization Measure}

\begin{figure}[t]
    \begin{subfigure}[b]{0.5\textwidth}
        \includegraphics[width=\textwidth]{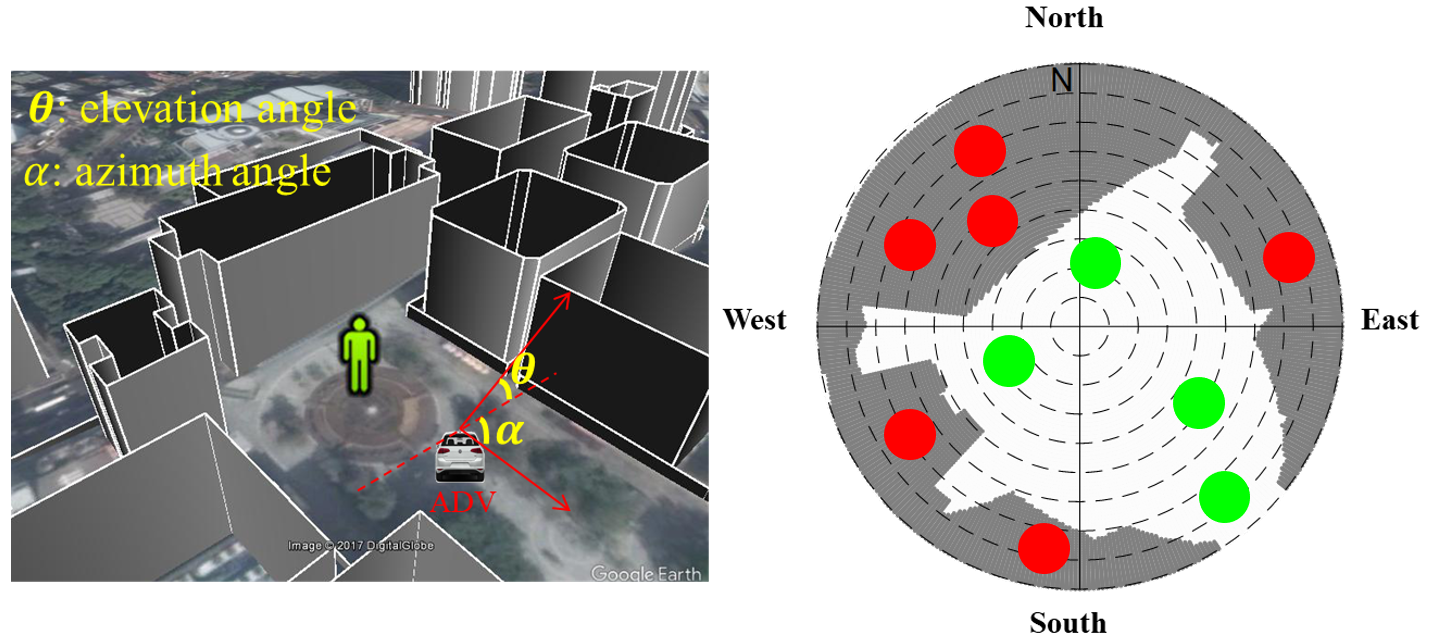}
        \caption[]%
        {{\small Urbanization Measure Definition}}    
    \end{subfigure}
    \hfill
    \begin{subfigure}[b]{0.5\textwidth}  
        \includegraphics[width=\textwidth]{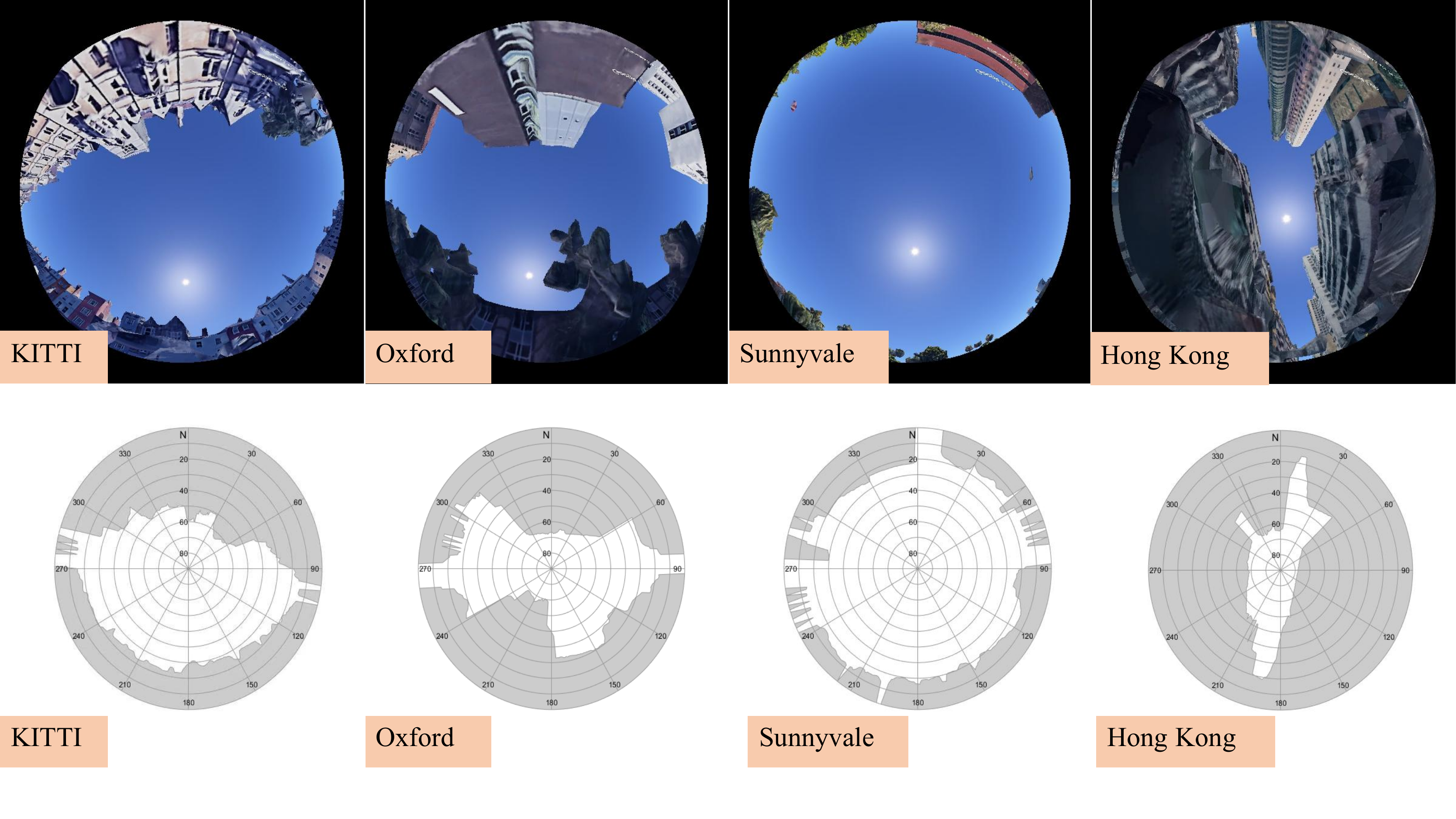}
        \caption[]%
        {{\small Skymask Comparison}}    
    \end{subfigure}
    \caption[ ]
    {\small Skymask Illustration. (a) We define the urbanization measure for localization tasks. Red dots are the satellites blocked by high-rise structures, and green dots denote satellites that are with in the line of sight. (b) A comparison of Skymasks in existing datasets.} 

\label{fig:long}
\label{fig:onecol}
\vspace{-2mm}
\end{figure}

In this section, we introduce our urbanization measure for localization challenges in different scenes. This metric is further used to quantitatively compare our dataset with other publicly available ones. 

While GNSS is an intuitive and cost-effective solution for localization challenges, its performance deteriorates in dense urban areas. One major reason for the poor performance is the high-rise structures in the urban landscapes: skyscrapers in a city could block satellite signals, resulting in a limited number of visible satellites for positioning. Moreover, the GNSS signal can be reflected by the building surface in urban canyon, introducing an extra travelling path in the GNSS time measurement. The reflected GNSS signal would induce the multipath effect and non-line-of-sight (NLOS) reception. More importantly, high-rise structures are often associated with dense population and heavy traffics in a city. Thus, it is intuitive to quantify the urbanization rate for localization problems based on the building structures near the ego vehicle. 

To define the urbanization rate of a specific location, a straightforward approach is to use Skymask, a polar plot of the structures' silhouette. Built upon a Skyplot \cite{skyplot}, the Skymask includes the building structures as a mask for satellites. As shown in Figure 2, the gray section denotes the sky-view blocked by the structures, and the white section denotes the clear overhead sky. One way to generate a Skymask is to employ the 3D building models which are widely used in the GNSS field \cite{3d-bld-1}\cite{3d-bld-2}\cite{3d-bld-3}. Another method is to utilize the fish eye camera on the top of the vehicle (shown in Figure 3(a)) and algorithms like normalized cut \cite{ncut} for image segmentation. 

To quantitatively analyze a Skymask, we further define two parameters: the mean mask elevation angle {$\mu_{MEA}$}, and the mask elevation angle standard deviation {$\sigma^2_{MEA}$}. The definitions are:

\[ \mu_{MEA}=\frac{\sum_{\alpha=1}^{N} \theta_{\alpha}}{N} \]
\[ \sigma^2_{MEA}=\sqrt{   \frac{\sum (\theta_{\alpha}-\mu_{MEA})^2}{N-1}} \]

\noindent where {$\theta_\alpha$} represents the elevation angle, which is highly correlated to the building heights, at a given azimuth angle {$\alpha$}, and {$N$} denotes the number of equally spaced azimuth angles from the Skymask. We usually use {$N$}=360, meaning the azimuth angle has a resolution of 1 degree. 

When the ego vehicle is in a dense urban area, the Skymask is usually dominated by high-rise structures, resulting in a large {$\mu_{MEA}$}, and a relatively small {$\sigma^2_{MEA}$}. In rural areas, on the other hand, both {$\mu_{MEA}$} and {$\sigma^2_{MEA}$} would be relatively small. In places with mixed high-rise and low-lying buildings, {$\sigma^2_{MEA}$} would be of a relatively large value. 

With the aforementioned urbanization measure, we evaluated the urbanization rate for current mapping/localization dataset, and the result is shown in Table 2 and Figure 2(b). Clearly, none of the three datasets matches level of urbanization in Hong Kong. The Skymask is generated by 3D building models in these area, and we developed Figure 2(b) through post-processing. 

\begin{table}[ht!]
\begin{center}
\caption{Quantified Urbanization Rate}
\vspace{2mm}
\begin{tabular}{|c|c|c|}
\hline
DATASET &   \multicolumn{2}{c|}{Urbanization Rate}  \\
\cline{2-3}
 NAME  &{$\mu_{MEA}$ (degree)} & {$\sigma^2_{MEA}$}  \\
\hline
\hline
KITTI \cite{KITTI} &  32.8 & {$21.4^2$}\\
Oxford RoboCar \cite{RobotCarDatasetIJRR}&  42.3 & {$16.3^2$}\\
Waymo (Sunnyvale) \cite{waymo_open_dataset}&  15.2 & {$7.9^2$}\\
\hline
Ours &  60.9 & {$15.9^2$}\\
\hline
\end{tabular}
\end{center}
\vspace{-8mm}
\end{table}

\section{The UrbanLoco Dataset}
Compared with the current public datasets for autonomous driving, our presented dataset covers more challenging urban trajectories with a full-sensor suite. The data was collected in densely populated areas in Hong Kong and San Francisco. These two cities have drastically different landscapes, driving behaviors (directions), architectures, and infrastructures. The trajectories cover urban canyons, tunnels, bridges, hills, sharp maneuvers, and other challenging scenes for the aforementioned mapping and localization solutions. More importantly, the scenes were filled with pedestrians, vehicles, trolleys, and cyclists. The sensors used were 1 LIDAR, 6 cameras (San Francisco), 1 fish eye camera (Hong Kong), 1 IMU, 1 GNSS receiver. The ground truth for both cities is given by the Novatel SPAN-CPT, a navigation system incorporates Real Time Kinematic (RTK) corrected GNSS signal and IMU measurements. In this section, we will explain our data collection platform and calibration in detail. 

\begin{figure}[t]
    \begin{subfigure}[b]{0.5\textwidth}
        \includegraphics[width=\textwidth]{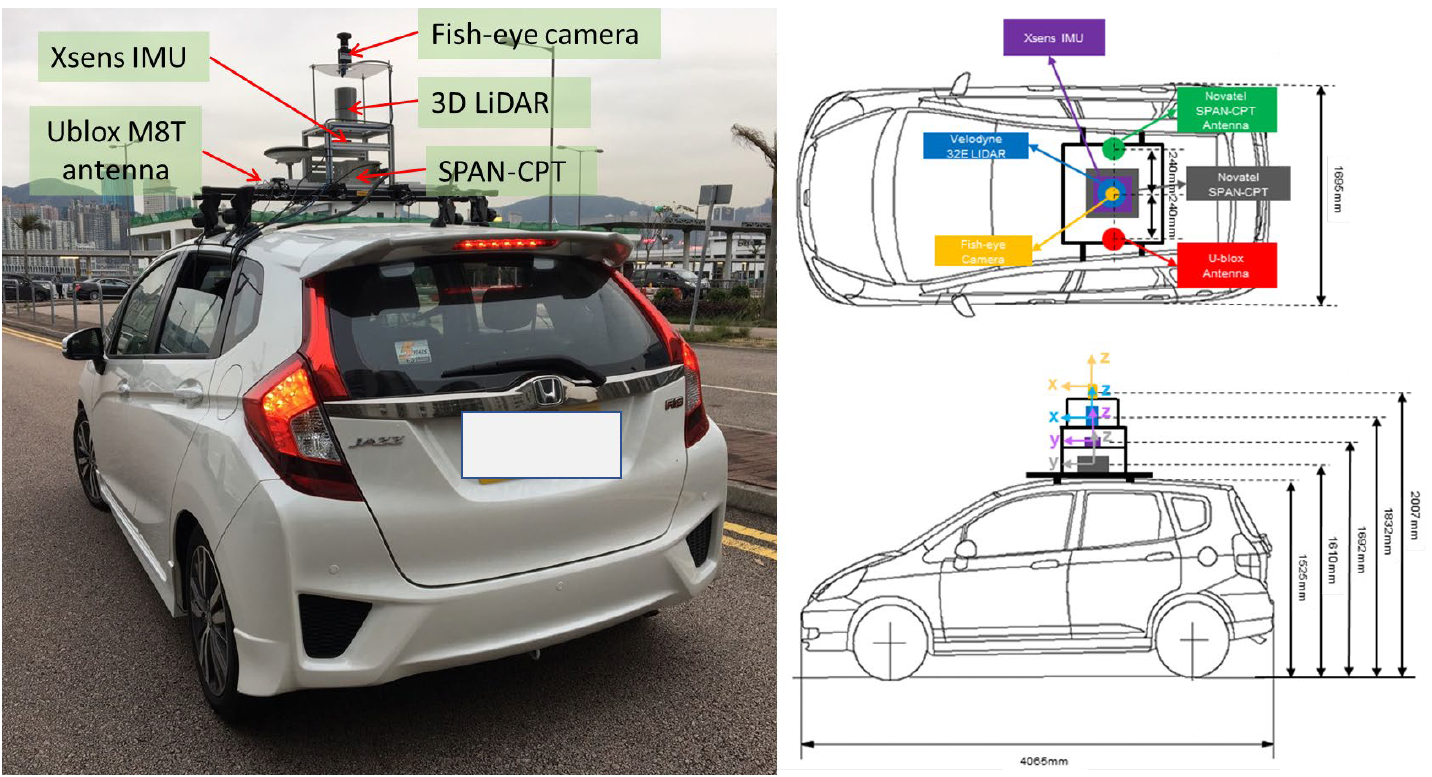}
        \caption[]%
        {{\small Hong Kong Data Collection Vehicle}}    
    \end{subfigure}
    \hfill
    \begin{subfigure}[b]{0.5\textwidth}  
        \includegraphics[width=\textwidth]{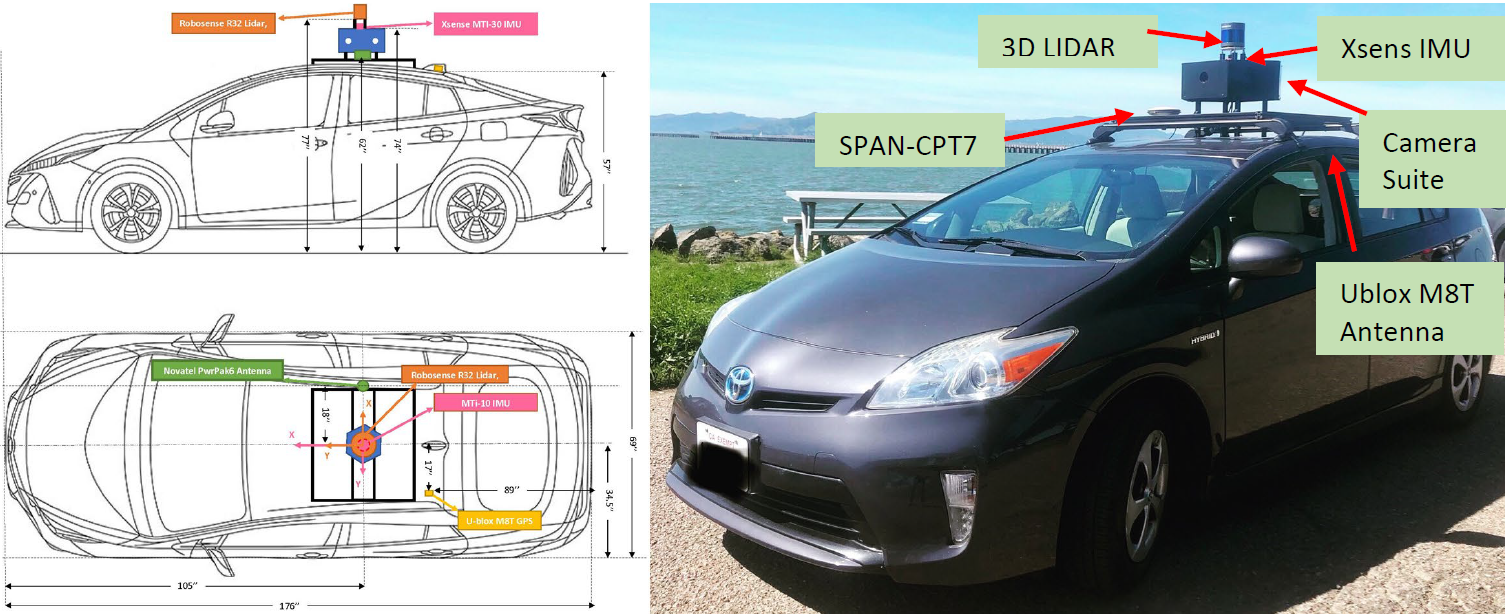}
        \caption[]%
        {{\small San Francisco Data Collection Vehicle}}    
    \end{subfigure}
    \caption[ ]
    {\small Data Collection Platforms. More detailed extrinsic parameters can be found on the data set website} 

\label{fig:long}
\label{fig:onecol}
\vspace{-2mm}
\end{figure}

\begin{table*}[ht!]
\begin{center}
\caption{Map Evaluation Results}
\vspace{2mm}
\begin{tabular}{|c|c|c|c|c|c|c|c|c|c|}
\hline
DATA  & Description  & Length & Method & \multicolumn{3}{c|}{Mean Translation Error (m)} &  \multicolumn{3}{c|}{Mean Rotation Error (deg)}  \\
\cline{5-10}
ROSBAG & & (km)& &X& Y& Z& Z&Y&X\\
\hline
\hline
& & & LOAM & 36.91 & 95.06 & 149.97 & 10.15 & 0.97 & 0.50\\
CA\_20190828155828 & Traffic, Structures & 5.9 & VINS & 42.68 & 42.68 & 23.24 & 6.04 & 0.69 & 2.05 \\
& & &  NDT & 184.05 & 94.01 & 337.91 & 3.68 & 4.43 & 5.89\\
\hline
& & & LOAM & 81.74 & 85.69 & 80.91 & 20.149 & 6.11 & 2.65\\
CA\_20190828173350 & Traffic, Hills & 3.2 & VINS & 26.745 & 34.756 & 14.83 & 8.12 & 0.81 & 2.47 \\
& & &  NDT & 129.27 & 71.13 & 253.76 & 10.10 & 0.76 & 11.07\\
\hline
& & & LOAM & 32.90 & 25.01 & 19.83 & 0.42 & 0.97 & 0.43\\
CA\_20190828184706 & Hills & 1.8 & VINS & 32.52 & 31.47 & 26.89 & 2.65 & 0.89 & 1.28 \\
& & &  NDT & 40.35 & 24.82 & 77.74 & 0.31 & 3.46 & 5.02\\
\hline
& & & LOAM & 29.53 & 21.97 & 15.50 & 12.29 & 1.58 & 2.85\\
CA\_20190828190411 & Hills, Maneuvers & 1.0 & VINS & 22.27 & 17.57 & 23.84 & 11.32 & 4.12 & 4.31 \\
& & &  NDT & 29.51 & 16.78 & 14.62 & 9.04 & 2.08 & 0.42\\
\hline
HK\_20190426101600 & Structures, Traffic & 0.8 & LOAM & 10.56 & 9.73 & 0.36 & 1.23 & 0.05 & 0.03 \\
& & &  NDT & 6.14 & 7.77 & 2.01 & 7.23 & 0.43 & 0.19\\
\hline
HK\_20190426100200 & Structures, Traffic & 0.7 & LOAM & 19.65 & 16.15 & 0.88 & 1.64 & 0.03 & 0.68 \\
& & &  NDT & 6.63 & 8.28 & 2.10 & 7.80 & 0.32 & 0.90\\
\hline
\end{tabular}
\end{center}
\vspace{-2mm}
\end{table*}

The platform for data collection in Hong Kong is a Honda Fit. The corresponding illustration and calibration coordinates are shown in Figure 3(a). All the localization-related sensors are equipped in a compact sensor kit on the top of the vehicle:

\setlist{nolistsep}
\begin{itemize}[noitemsep]
    \item \textbf{LIDAR} Velodyne HDL 32E, 360 Horizontal Field of View (FOV), -30~+10 vertical FOV, 80 meters in range, 10 Hz;
    \item \textbf{Camera} Grasshopper3 5.0 MP (GS3-U3-51S5C-C), fish eye lens Fujinon FE185C057HA-1, 185 HFOV, 185 V-FOV, 10 Hz;
    \item \textbf{IMU} Xsens Mti 10, 100 Hz;
    \item \textbf{GNSS} Ublox M8T, GPS/BeiDou, 1Hz;
\end{itemize}

We used Ublox M8T to collect raw GNSS signals to enable future researches on GNSS positioning: e.g., correcting the NLOS measurements \cite{ww18}. A sky-pointing fish eye camera is also applied to capture the sky view. With the camera, it is possible to quantify satellite visibility based on the sky view image to monitor and to improve the quality of GNSS positioning \cite{kato2016nlos}  \cite{ww19}. The 3D LIDAR is employed to scan the environment for HD Map generation and SLAM \cite{Grisetti} purpose.

Lastly, the Xsens IMU is employed to collect raw acceleration and orientation measurements at high frequency. 

The platform for data collection in California is a Toyota Prius as illustrated in Figure 3(b). Slightly different from the Hong Kong platform, the following sensors are used in California:

\setlist{nolistsep}
\begin{itemize}[noitemsep]
    \item \textbf{LIDAR} RS-LIDAR-32, 360 Horizontal Field of View (FOV), -25~+15 vertical FOV, 80 meters in range, 10 Hz;
    \item \textbf{Camera} Six FLIR Blackfly S USB3, 2048*1536, 10 degree overlap on each side, 10Hz;
    \item \textbf{IMU} Xsens Mti 10, 100 Hz;
    \item \textbf{GNSS} Ublox M8T, GPS/GLONASS, 1Hz
\end{itemize}

The added six 360-degree view cameras are synchronized and calibrated with the LIDAR. These two kinds of sensors are synchronized via triggering the front-camera when the rotating LIDAR beam passes the front center line (software trigger), and the rest cameras are triggered sequentially at a fraction of the running LIDAR frequency. To compensate the possible delay in triggering signal, the triggering time is adjusted to align LIDAR and cameras. Thus the specific part of the LIDAR scan and the corresponding camera image contain information at the same wall time. The intrinsic matrices of the cameras are calibrated through the kalibr toolbox \cite{kalibr}, and the extrinsic calibration matrices are solved through Autoware \cite{autoware}. 

The ground truth used in this dataset is provided by Novatel SPAN-CPT, a GNSS-IMU navigation system. The device is widely used for accurate localization assignments on mobile platforms. For the GNSS receiver, the received signal is corrected from the RTK signal sent from local public base stations. When the satellite visibility is satisfactory, the ground truth error is within 2 cm. In cases of a complete loss of the GNSS signal, the device is still able to output a continuous trajectory based on IMU measurements. The calibration certificate dictates that the error after 10s of GNSS blackout is within 12 cm.

\section{Evaluation}

\begin{figure}[t]
        \begin{subfigure}[b]{0.235\textwidth}
            \includegraphics[width=\textwidth]{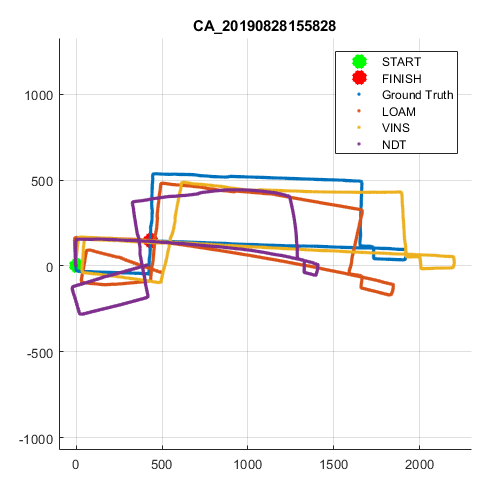}
            \caption[]%
            {{\scriptsize CA\_20190828155828 }}    
        \end{subfigure}
        \hfill
        \begin{subfigure}[b]{0.235\textwidth}  
            \includegraphics[width=\textwidth]{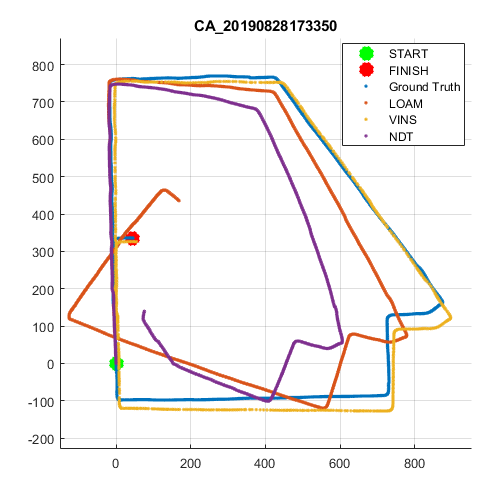}
            \caption[]%
            {{\scriptsize CA\_20190828173350}}    
        \end{subfigure}
        \begin{subfigure}[b]{0.235\textwidth}   
            \includegraphics[width=\textwidth]{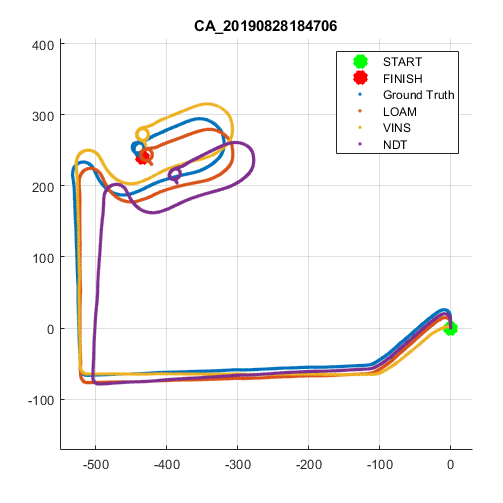}
            \caption[]%
            {{\scriptsize CA\_20190828184706 }}  
        \end{subfigure}
        \hfill
        \begin{subfigure}[b]{0.235\textwidth}   
            \includegraphics[width=\textwidth]{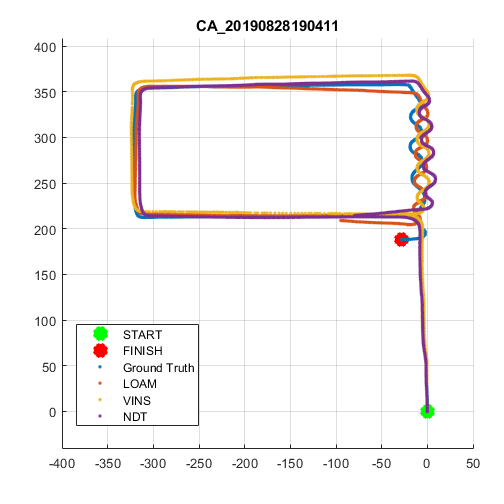}
            \caption[]%
            {{\scriptsize CA\_20190828190411}}    
        \end{subfigure}
        \begin{subfigure}[b]{0.235\textwidth}   
            \includegraphics[width=\textwidth]{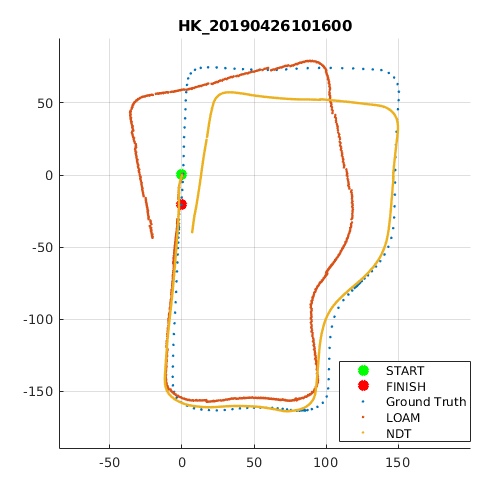}
            \caption[]%
            {{\scriptsize HK\_20190426100200 }}  
        \end{subfigure}
        \hfill
        \begin{subfigure}[b]{0.235\textwidth}   
            \includegraphics[width=\textwidth]{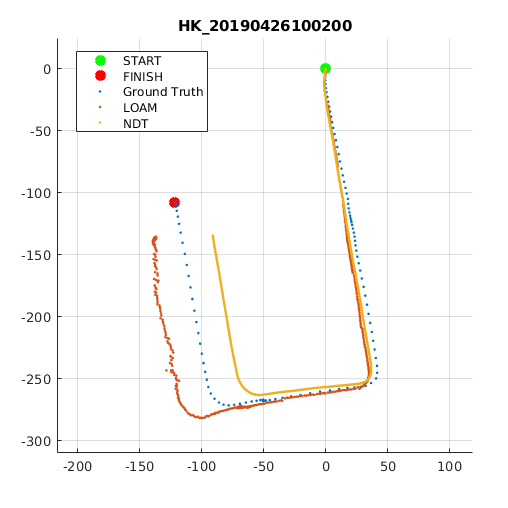}
            \caption[]%
            {{\scriptsize HK\_20190426101600}}    
        \end{subfigure}
        \caption[]
        {\small Map Evaluation Results} 
    \label{fig:long}
    \label{fig:onecol}
    \vspace{-2mm}
\end{figure}

A few state of the art SLAM algorithms were evaluated with our dataset. For the tested open-source LIDAR and camera based methods, the results were less satisfactory in the our highly urbanized scenarios. A quantitative evaluation is given in Table 3, and a collection of 2D trajectory plots are shown in Figure 4. The dynamic objects in the moving scene largely contributed to the failures of SLAM algorithms, as most of these algorithms use static feature points to estimate the pose of the vehicle. Map $CA\_20190928173350$, collected during the rush hour on the one of the busiest street in San Francisco, is the most distorted map produced with current SLAM algorithms. 


\subsection{Laser-based Methods}

The tested LIDAR-based methods were LOAM \cite{LOAM} (Ranking No.2 on KITTI \cite{KITTI}), and NDT \cite{NDT} (A similar algorithm ranked No.24 on KITTI \cite{KITTI}). For LOAM, the KITTI reported average translation error was less than 5.7m/km, and the rotation error was 0.0013 degree/m. However, when retesting the algorithm with the open source LOAM algorithm by Leonid Laboshin \cite{loam_code}, the translation performance deteriorated significantly. For cases filled with dynamic objects (Figure 4(e,f)), the performance dropped to above 10m/km on the horizontal directions. In cases of drastic altitude changes (Figure 4(b,c,d)), the translation performance further decreased to over 20m/km. As for rotation, the rotation performance decreased when the vehicle experienced sharp maneuvers (Figure 4(c,d)). 

For NDT \cite{NDT} family solutions, the best algorithm on KITTI reached 8.9m/km in translation error and 0.003 degree/m in rotation. In our evaluation, we applied the package prepared by Kenji Koide \cite{hdl}, and fine-tuned the parameters three times for the best performance. Furthermore, half real-time playback rate was used to guarantee solution convergence. However, the result is less satisfactory: in the longest testing route (5.9km, FIgure 4(a)), the translation error was more than 100 meters overall. It was also observed that the algorithm performed poorly on altitude estimation. 

\subsection{Vision-based Methods}
As for visual odometer estimation, we used the open-source VINS-MONO \cite{qin2017vins} algorithm developed by Qin et al. VINS-MONO tightly couples visual odometry with IMU estimation to output an optimized localization result. As pointed out in \cite{qin2017vins}, the algorithms out-performed most existing visual odometry methods. During the experiment, we noticed that the visual odometry was very sensitive to changes in light conditions: the algorithm failed when entering-exiting a tunnel. After using half-real time playback rate, we generated the continuous trajectory successfully. For performance, while the algorithm slightly out-performed the LIDAR based method in translation, the error in roll/pitch/yaw angle estimation was worse. The performance further deteriorated when the path is filled with sharp maneuvers (Figure 4(d)). Since VINS-MONO takes time for initialization, a rigid body transformation \cite{rigidTF} was applied to the constructed map before evaluation. 

From the aforementioned evaluation and analysis, it is safe to say that our dataset in Hong Kong and San Francisco addresses urban driving challenges adequately. Indeed, the dataset is a valuable test field for future urban-focused localization solutions.


\section{Conclusion}
We have presented a challenging full sensor suite dataset for autonomous vehicle mapping and localization. Comparing with current datasets, our dataset contains trajectories from more challenging scenes, and a few state of the art algorithms performed poorly on our dataset. We also defined the urbanization measure to further quantify the challenges in different scenarios. More importantly, our sensor suite contains various sensors for mapping and localization purposes: LIDAR, cameras, IMU, and GNSS receivers. In the future, additional data would be added to this dataset. We would also exploit the possibilities of presenting benchmarks similar to the KITTI\cite{KITTI} dataset.

\section*{Acknowledgement}
Authors hereby thank supports from Robosense, whose R32 LIDAR donation is critical for data acquisition. We also thank Di Wang for his contributions on vehicle instrumentation.

{\small
\bibliographystyle{ieee}
\bibliography{egbib}
}

\end{document}